\documentclass{article}

\usepackage[colorlinks,bookmarksnumbered,citecolor=orange,urlcolor=orange]{hyperref}
\usepackage{color, colortbl}
\usepackage{graphicx}
\DeclareGraphicsExtensions{.pdf,.png}
\usepackage{amsmath}
\usepackage{amssymb}
\usepackage{bbm}
\usepackage{algorithm}
\usepackage[noend]{algpseudocode}
\usepackage{wrapfig}
\usepackage{multirow}
\usepackage{booktabs}
\usepackage{enumitem}
\usepackage{xcolor}
\usepackage{authblk}

\usepackage[final]{corl_2020} 

\usepackage{titlesec}
\titlespacing*{\subsection}{0pt}{0.1\baselineskip}{0.05\baselineskip}
\titlespacing*{\section}{0pt}{0.2\baselineskip}{0.1\baselineskip}

\title{Learning Latent Representations \\to Influence Multi-Agent Interaction}

\makeatletter
\renewcommand\AB@affilsepx{, \protect\Affilfont}
\makeatother
\author[1]{Annie Xie}
\author[2]{Dylan P. Losey}
\author[1]{Ryan Tolsma}
\author[1]{Chelsea Finn}
\author[1]{Dorsa Sadigh}
\affil[1]{Stanford University}
\affil[2]{Virginia Tech}

\begin{document}
\maketitle


\begin{abstract}
Seamlessly interacting with humans or robots is hard because these agents are non-stationary. They update their policy in response to the ego agent's behavior, and the ego agent must anticipate these changes to co-adapt. Inspired by humans, we recognize that robots do not need to explicitly model every low-level action another agent will make; instead, we can capture the \textit{latent strategy} of other agents through high-level representations. We propose a reinforcement learning-based framework for learning latent representations of an agent's policy, where the ego agent identifies the relationship between its behavior and the other agent's future strategy. The ego agent then leverages these latent dynamics to influence the other agent, purposely guiding them towards policies suitable for co-adaptation. Across several simulated domains and a real-world air hockey game, our approach outperforms the alternatives and learns to influence the other agent\footnote{Videos of our results are available at our project webpage:  \href{https://sites.google.com/view/latent-strategies/}{\textcolor{orange}{https://sites.google.com/view/latent-strategies/}}.}.
\end{abstract}



\section{Introduction}
\label{sec:introduction}

Although robot learning has made significant advances, most algorithms are designed for robots acting in isolation. In practice, interaction with humans and other learning agents is inevitable. Such interaction presents a significant challenge for robot learning: the other agents will \textit{update} their behavior in response to the robot, continually changing the robot's learning environment. For example, imagine an autonomous car that is learning to regulate the speed of nearby human-driven cars so that it can safely slow those cars if there is traffic or construction ahead. The first time you encounter this autonomous car, your intention is to act cautiously: when the autonomous car slows, you also slow down. But when it slowed without any apparent reason, your strategy changes: the next time you interact with this autonomous car, you drive around it without any hesitation. The autonomous car --- which originally thought it had learned the right policy to slow your car --- is left confused and defeated by your updated behavior!

To seamlessly \textit{co-adapt} alongside other agents, the robot must anticipate how the behavior of these other agents will change, and model how its own actions affect the hidden intentions of others. Prior works that study these interactions make restrictive assumptions about the other agents: treating the other agents as stationary~\cite{davison1998predicting,stone2000defining,mealing2015opponent,albrecht2015hba,grover2018learning}, sharing a learning procedure across other agents~\cite{foerster2016learning,lowe2017multi,foerster2018counterfactual}, directly accessing the underlying intentions and actions of other agents~\cite{albrecht2015hba,he2016opponent,hernandez2017efficiently, grover2018learning, foerster2018learning, everett2018learning}, or simplifying the space of intentions with hand-crafted features~\cite{kelley2008understanding,bandyopadhyay2013intention,bai2015intention,sadigh2016information,basu2019active, losey2019robots}.

\begin{figure}
\includegraphics[width=.95\columnwidth]{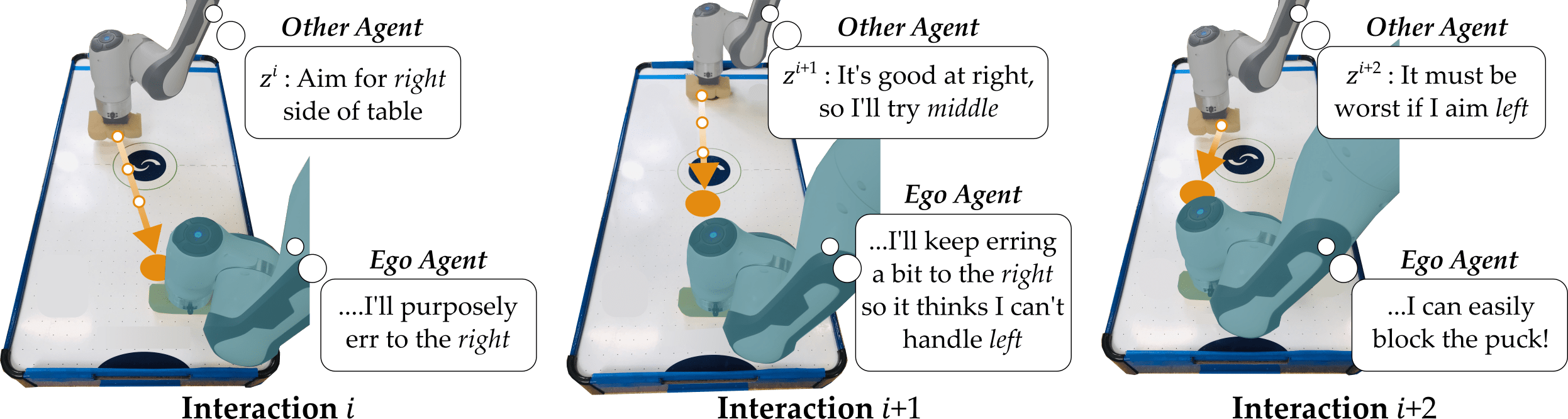}
\centering
\vspace{-0.5em}
\caption{\small{Ego agent learning to play hockey against a non-stationary robot. The other robot updates its policy between interactions to exploit the ego agent's weaknesses. Over repeated interactions, the ego agent learns to represent each opponent policy as a high-level latent strategy, $z$, and also recognizes that the opponent updates $z$ to aim away from where the ego agent last blocked. The ego agent leverages these latent dynamics to \textit{influence} the other robot, learning a policy that guides the opponent into aiming where the ego agent can block best.}}
\label{fig:front}
\vspace{-2em}
\end{figure}

By contrast, we focus on general settings where the ego agent (e.g., a robot) repeatedly interacts with non-stationary, separately-controlled, and partially-observable agents. People seamlessly deal with these scenarios on a daily basis (e.g., driving, walking), and they do so without explicitly modeling every low-level aspect of each other's policy \cite{rubinstein1998modeling}. Inspired by humans, we recognize that:
\begin{center}
    \vspace{-0.5em}
    \emph{The ego agent only observes the low-level actions of another agent, but --- just as in human-human interaction --- it is often sufficient to maintain a high-level policy representation. We refer to this representation as the \emph{latent strategy}, and recognize that latent strategies can change over time.}
    \vspace{-0.5em}
\end{center}
Learning the other agent's strategy allows the robot to \textit{anticipate} how they will respond during the current interaction. Furthermore, modeling how the latent strategy changes over time --- i.e., the \textit{latent dynamics} --- enables the robot to predict how its own behavior will \textit{influence} the other agent's future intentions (see Fig.~\ref{fig:front}). Overall, we leverage latent representations of the other agent's policy to make the following contributions towards learning in non-stationary multi-agent interaction:

\textbf{Learning Latent Representations.} We introduce an RL-based framework for multi-agent interaction that learns both latent strategies and policies for responding to these strategies. Our framework implicitly models how another agent's strategy changes in response to the ego agent's behavior.

\textbf{Influencing Other Agents.} Ego agents leveraging our proposed approach are encouraged to purposely change their policy to influence the latent strategies of other agents, particularly when some strategies are better suited for co-adaptation than other strategies.

\textbf{Testing in Multi-Agent Settings.} We compare our approach to state-of-the-art methods across four simulated environments and a real robot experiment, where two $7$-DoF robot arms play air hockey. Our approach outperforms the alternatives, and also learns to influence the other agent.

\section{Related Work}
\label{sec:related_work}

\textbf{Opponent Modeling.} Several prior works in multi-agent RL (MARL) and human-robot interaction (HRI) handle non-stationary interactions by \textit{modeling} the other agents. These approaches either model intention~\cite{wang2013probabilistic,raileanu2018modeling}, assume an assignment of roles~\cite{losey2020learning}, exploit opponent learning dynamics~\cite{foerster2018learning, zhang2010multi}, or incorporate features of opponent behavior, which can be modeled using a recursive reasoning paradigm~\cite{baker2011bayesian}, handcrafted~\cite{he2016opponent}, or
learned~\cite{grover2018learning, rabinowitz2018machine, papoudakis2020variational}.
Since explicitly modeling and reasoning over the opponent's intentions or policy can quickly become computationally intractable, we sidestep this recursion by learning a low-dimensional representation of the other agent's behavior. Unlike prior works with learned models, however, we recognize that this representation can dynamically change and be influenced by the ego agent's policy.

\textbf{Multi-Agent RL.} Alternate approaches have adopted a \textit{centralized} training framework~\cite{lowe2017multi,foerster2018counterfactual,woodward2020learning} or learned \textit{communication} protocols between agents~\cite{lazaridou2016multi,foerster2016learning,sukhbaatar2016learning,mordatch2018emergence}. We do not require either centralization or communication: hence, our approach can operate in more general multi-agent systems, such as those with humans or decentralized agents. Notably, our setting does make a different assumption --- that other agents have predictable strategies which can be captured by latent representations. Although this is not always the case (e.g., when other agents are RL-based), our assumption is reasonable in a wide range of scenarios, especially when interacting with humans~\cite{rubinstein1998modeling}.

\textbf{Influence Through Interactions.} Influential behaviors emerge in MARL by directly shaping other agents' policy updates~\cite{foerster2018learning} and maximizing the mutual information between agents' actions~\cite{jaques2019social}. While these works encourage \textit{influencing} by modifying the learning objective, e.g., through auxiliary rewards~\cite{jaques2019social}, in our approach influence arises without any explicit reward or encouragement. 
Within HRI, most prior work studies influence in the context of a \emph{single} interaction.
In particular, robots have demonstrated influence over humans by leveraging human trust~\cite{losey2019robots,chen2018planning}, generating legible motions~\cite{dragan2013legibility}, and identifying and manipulating structures that underlie human-robot teams~\cite{kwon2019influencing}. 
Other works model the effects of robot actions on human behavior in driving~\cite{sadigh2016planning,sadigh2018planning} and handover~\cite{bestick2016implicitly} scenarios in order to learn behavior that influences humans. 
Unlike these works, we learn to influence without recovering or accessing the other agent's reward function.

\textbf{Partial Observability in RL.} We can view our setting as an instantiation of the partially observable Markov Decision Process (POMDP)~\cite{kaelbling1998planning}, where the other agent's hidden strategy is the latent state. Approximate POMDP solutions based on representation learning have scaled to high-dimensional state and action spaces~\cite{igl2018deep} such as image observations~\cite{hafner2019learning,lee2019stochastic}. However, we assume the other agent's latent strategy is \textit{constant} throughout an episode, and then \textit{changes} between episodes: we exploit this structure by modeling the ego agent's environment as a sequence of hidden parameter MDPs (HiP-MDP)~\cite{doshi2016hidden}. Recent work~\cite{xie2020deep} similarly models non-stationary environments as a series of HiP-MDPs, but assumes that the hidden parameters evolve \textit{independently}. Unlike this work, we recognize that the ego agent's behavior can also influence the other agent's latent strategy.

\textbf{Robotic Air Hockey.} Finally, prior works have also developed robots that play air hockey, focusing on vision-based control for high-speed manipulation~\cite{bishop1999vision,wang2002vision,bentivegna2002humanoid,ogawa2011development}. Motivated by a similar objective to ours,~\citet{namiki2013hierarchical} hand-design a system that switches strategies based on the opponent. However, we aim to autonomously learn both the opponent's strategies and how to respond to them.

\section{Repeated Interactions with Non-Stationary Agents}
\label{sec:problem_statement}

In this section we formalize our problem statement. Although our approach will extend to situations with $N$ other agents, here we focus on \textit{dyads} composed of the ego agent and one other agent. This other agent could be an opponent in competitive settings or a partner in collaborative settings.

Recall our motivating example, where an autonomous car (the ego agent) is learning to regulate the speed of a human-driven car (the other agent) for safety in heavy traffic or construction zones. This specific autonomous car and human-driven car have a similar commute, and encounter each other on a daily basis. Over these \textit{repeated interactions}, the human continually updates their policy: e.g., shifting from following to avoiding, and then back to following after the autonomous car establishes trust. The autonomous car has access to the history of interactions, including being able to sense the human-driven car's speed and steering angles. Inspired by our insight, the autonomous car assumes that these low-level actions are the result of some \emph{high-level intention} sufficient for coordination: the human driver's policy is therefore captured by the latent strategy $z$. During each interaction, $z$ is constant --- e.g., the human's strategy is to follow the autonomous car --- but $z$ changes between interactions according to the human driver's latent dynamics. Importantly, the ego agent never explicitly observes the hidden intentions of the other agent, and therefore must learn both the latent strategies and their latent dynamics from local, low-level observations.

\textbf{Strategy Affects Transitions and Rewards During Interaction.} Let $i$ index the current interaction, and let latent strategy $z^i$ be the low-dimensional representation of the other agent's policy during the $i$-th interaction. We formulate this interaction as a hidden parameter Markov decision process (HiP-MDP) \cite{doshi2016hidden}. Here the HiP-MDP is a tuple $\mathcal{M}^i = \langle \mathcal{S}, \mathcal{A}, \mathcal{Z}, \mathcal{T}, R, H \rangle$, where $s \in \mathcal{S}$ is the state of the ego agent and $a \in \mathcal{A}$ is the ego agent's action. From the ego agent's perspective, the other agent's policy affects the environment: if the human's strategy is to avoid the autonomous car, this may induce other cars to also pass (altering the system dynamics), or prevent the autonomous car from successfully blocking (changing the reward). Accordingly, we let the unknown transition function $\mathcal{T}(s' ~|~ s, a, z^i)$ and the reward function $R(s, z^i)$ depend upon the latent strategy $z^i \in \mathcal{Z}$. The interaction ends after a fixed number of timesteps $H$. Over the entire interaction, the ego agent experiences the trajectory $\tau^i = \{(s_1, a_1, r_1), \ldots, (s_H, a_H, r_H)\}$ of states, actions, and rewards.

\textbf{Strategy Changes Between Interactions.} In response to the ego agent's behavior for the $i$-th interaction, the other agent may update its policy for interaction $i+1$. For example, imagine following an autonomous car which brakes irrationally. The next time you encounter this car, it would be natural to switch lanes, speed up, and move away! We capture these differences in low-level policy through changes in the high-level strategy, such that $z$ has Markovian latent dynamics: $z^{i+1} \sim f( \cdot ~|~ z^i, \tau^i)$.

\textbf{Strategies are Influenced Across Repeated Interactions.} Combining our formalism both \textit{during} and \textit{between} interactions, we arrive at the complete problem formulation. Over repeated interactions, the ego agent encounters a series of HiP-MDPs
$(\mathcal{M}^1, \mathcal{M}^2, \mathcal{M}^3, \ldots)$, where the other agent has latent strategy $z^i$ in $\mathcal{M}^i$. The ego agent's objective is to maximize its cumulative discounted reward across interactions. Because the other agent changes its policy in response to the ego agent, our problem is not as simple as greedily maximizing rewards during each separate interaction. Instead, the ego agent should purposely \textit{influence} the other agent's strategy to achieve higher long-term rewards. For example, suddenly stopping during the current interaction may cause the human to immediately slow down, but in future interactions the human will likely avoid our autonomous car altogether! Thus, an intelligent agent should take actions that will lead to being able to slow the human when needed, such as in traffic or near construction zones.

\section{Learning and Influencing Latent Intent (LILI)}
\label{sec:method}

\begin{figure}
\includegraphics[width=1.0\columnwidth]{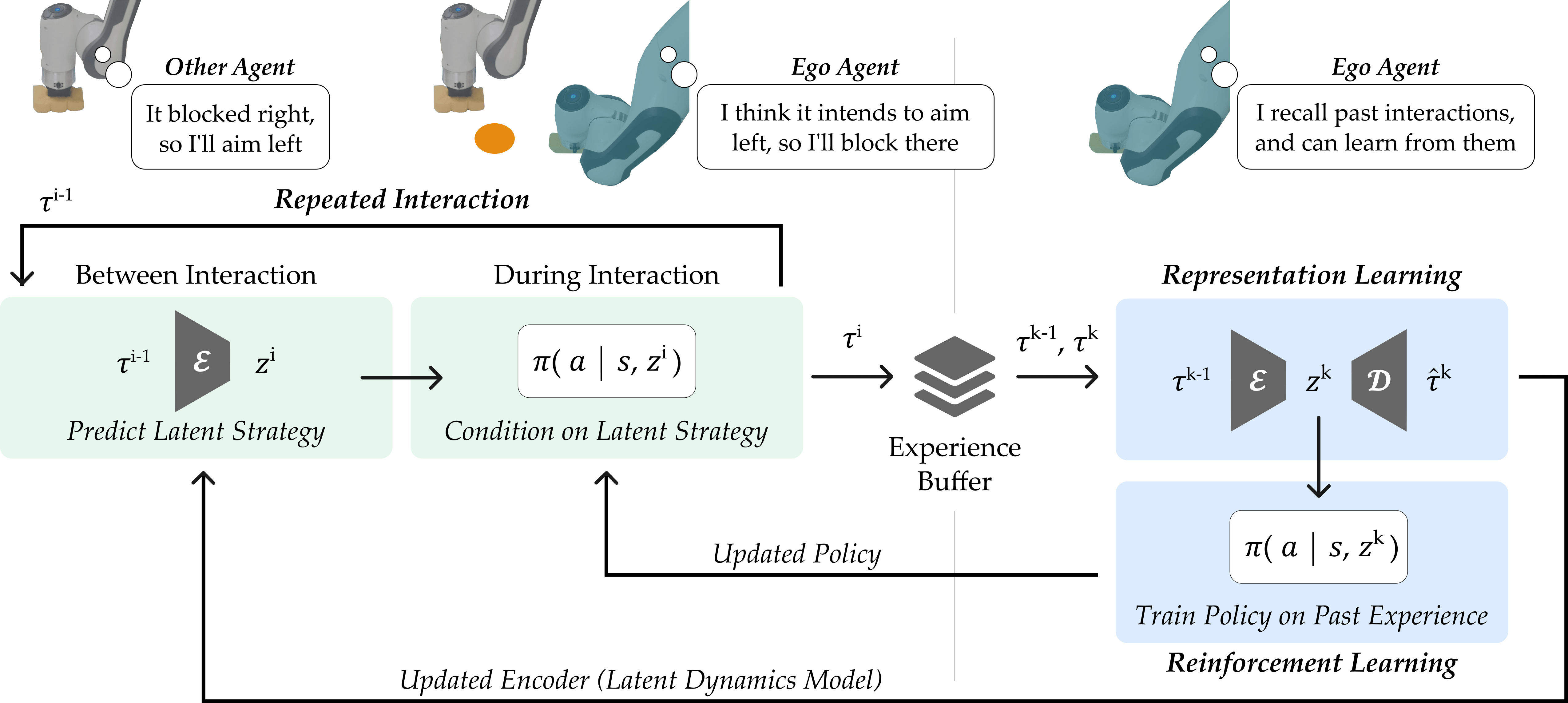}
\vspace{-1.5em}
\centering
\caption{\small{Our proposed approach for learning and leveraging latent intent. Left: Across repeated interactions, the ego agent uses their previous experience $\tau^{i-1}$ to predict the other agent's current latent strategy $z^i$, and then follows a policy $\pi$ conditioned on this prediction. Right: The ego agent learns by sampling a pair of consecutive interactions, and (a) training the encoder $\mathcal{E}$ and decoder $\mathcal{D}$ to correctly predict experience $\tau^k$ given $\tau^{k-1}$, while simultaneously (b) using model-free RL to maximize the ego agent's long-term reward.}}
\vspace{-1.0em}
\label{fig:method}
\end{figure}

In this section we present our proposed approach to learn and influence latent strategies in multi-agent interaction. Refer to Fig.~\ref{fig:method} for an overview: the ego agent simultaneously learns an encoder --- which captures the other agent's latent strategy from low-level observations --- and a policy --- which is conditioned on the inferred latent strategy. During each interaction, the robot predicts what the other agent's strategy \textit{will be} based on their last interaction, and reacts using its strategy-conditioned policy. In what follows, we explain the different aspects of this approach.

\subsection{Learning Latent Strategies from Local Observations}

Our first step is to learn to \textit{represent} the behavior of other agents. Recall that during the $i$-th interaction the ego agent experiences $\tau^i$, a trajectory of local states, actions, and rewards. We leverage this local experience $\tau^i$ to anticipate $z^{i+1}$, the latent strategy that the other agents will follow during the \textit{next} interaction. 
More specifically, we introduce an encoder $\mathcal{E}_\phi$ that embeds the experience $\tau^i$ and predicts the subsequent latent strategy $z^{i+1}$.
Hence, $\mathcal{E}_\phi$ models $f(z^{i+1}~|~\tau^i)$, an approximation of the latent dynamics $f(z^{i+1}~|~z^i, \tau^i)$. We use this approximation because the other agent's current strategy $z^i$ can often be inferred from the ego agent's experience $\tau^i$, and we find that this functional form works well across our experimental domains. However, the actual strategies of the other agents are never explicitly observed --- how can we learn our encoder without these labels?

To address this challenge, we recognize that the latent strategy of the other agent determines how they will react to the ego agent's behavior, which in turn determines the dynamics and reward functions experienced by the ego agent during the current interaction. We therefore decode the latent strategy $z^{i+1}$ using a decoder $\mathcal{D}_\psi$ that \textit{reconstructs} the transitions and rewards observed during interaction $i+1$. Given a sequence of interactions $\tau^{1:N}$, we train our encoder $\mathcal{E}_\phi$ and decoder $\mathcal{D}_\psi$ with the following maximum likelihood objective $\mathcal{J}_\text{rep}$:
\begin{equation} \label{eq:M1}
    \max_{\phi,\psi}~\sum_{i=2}^N \sum_{t=1}^H \log p_{\phi, \psi} (s_{t+1}^i, r_t^i ~|~ s_t^i, a_t^i, \tau^{i-1})
\end{equation}
where $\mathcal{E}_\phi$ produces a deterministic embedding $z^i$ given $\tau^{i-1}$, and $\mathcal{D}_\psi$ outputs a distribution over next state $s_{t+1}$ and reward $r_t$ given current state $s_t$, action $a_t$, and embedding $z^i$.
The encoder is particularly important here:
we can use $\mathcal{E}_\phi$ to anticipate the other agent's next strategy. 

\subsection{Reinforcement Learning with Latent Strategies}
Given a prediction of what latent strategy the other agent is following, the ego agent can intelligently \textit{react}. Imagine that the autonomous car in our example knows that the human's strategy is to pass if the autonomous car slows significantly, and the autonomous car is approaching heavy traffic. Under this belief, the autonomous car should slow gradually, keeping the human behind as long as possible to reduce their speed.  But if --- instead --- the agent thinks that the human will also slow down with it, the agent can immediately brake and regulate the human to a safe speed. In other words, different latent strategies necessitate different policies. We accordingly learn an ego agent policy $\pi_\theta(a ~|~ s, z^i)$, where the ego agent makes decisions conditioned on the latent strategy prediction $z^i = \mathcal{E}_\phi(\tau^{i-1})$.

\subsection{Influencing by Optimizing for Long-Term Rewards}

What we ultimately want is not for the ego agent to simply \textit{react} to the predicted latent strategy; instead, an intelligent agent should proactively \textit{influence} this strategy to maximize its reward over repeated interactions. The key to this proactive behavior is $\mathcal{E}_\phi$, the ego agent's approximation of the latent dynamics. As the ego agent learns $\mathcal{E}_\phi$, it can purposely alter the interaction $\tau^i$ to influence the other agents towards a desired $z^{i+1}$. Crucially, some latent strategies $z^{i+1}$ may be better than others.

Jumping back to our example, the autonomous car is rewarded for regulating the human-driven car's speed near heavy traffic and construction zones. The human's strategies $\{z_1, z_2\} \in \mathcal{Z}$ include avoiding and passing the autonomous car ($z_1$) or trusting and following the autonomous car ($z_2$). From the ego agent's perspective, $z_2$ is a better latent strategy: if the autonomous car guides the human towards this strategy, it can safely reduce the human's speed when necessary. To learn influential behavior, we train the ego agent policy $\pi_\theta$  to maximize rewards across multiple interactions:
\begin{equation} \label{eq:M3}
    \max_\theta~ \sum_{i=1}^{\infty} \gamma^i~ \mathbb{E}_{\rho_{\pi_\theta}^i} \left[ \sum_{t=1}^H  R(s, z^i) \right] 
\end{equation}
where $\gamma \in [0, 1)$ is a discount factor and $\rho_{\pi_\theta}^i(\tau)$ is the trajectory distribution induced by $\pi_\theta$ under dynamics $\mathcal{T}(s' | s, a, z^i)$. Maximizing this RL objective naturally motivates the ego agent to influence the other agent. Specifically, the ego agent learns to generate interactions $\tau^i$ which lead the other agent to adopt future latent strategies that the ego agent can exploit for higher rewards. Within our example, the ego agent purposely establishes trust with the speeding human, guiding them towards the advantageous $z_2$ before gradually reducing speed near traffic or construction.

\subsection{Implementation}
\begin{wrapfigure}{r}{0.6\textwidth}
    \begin{minipage}{0.6\textwidth}
        \vspace{-0.8cm}
        \begin{algorithm}[H]
            \caption{Learning and Influencing Latent Intent (LILI)}
            \label{alg:latent_strategies}
            \begin{algorithmic}[1]
            \Require Learning rates $\alpha_Q$, $\alpha_\pi$, $\alpha_\phi$, $\alpha_\psi$
            \State Randomly initialize $\theta_Q$, $\theta_\pi$, $\phi$, and $\psi$
            \State Initialize empty replay buffer $\mathcal{B}$
            \State Assign $z^1 \gets \vec{0} $
            \For {i = 1, 2, \dots}
                \State Collect interaction $\tau^i$ with $\pi_\theta(a | s, z^i)$
                \State Update replay buffer $\mathcal{B}[i] \gets \tau^i$
                \For {j = 1, 2, \dots, N}
                    \State Sample batch of interaction pairs $(\tau, \tau') \sim \mathcal{B}$
                    \State $\theta_Q \gets \theta_Q - \alpha_Q \nabla_{\theta_Q} \mathcal{J}_Q$ \Comment{Critic update}
                    \State $\theta_\pi \gets \theta_\pi - \alpha_\pi \nabla_{\theta_\pi} \mathcal{J}_\pi$ \Comment{Actor update}
                    \State $\phi \gets \phi - \alpha_\phi \nabla_{\phi} \left( \mathcal{J}_Q - \mathcal{J}_\text{rep} \right)$ \Comment{Encoder update}
                    \State $\psi \gets \psi + \alpha_\psi \nabla_\psi \mathcal{J}_\text{rep}$ \Comment{Decoder update}
                \EndFor
                \State Assign $z^{i+1} \gets \mathcal{E}_\phi(\tau^i)$
            \EndFor
            \end{algorithmic}
        \end{algorithm}
        \vspace{-0.8cm}
    \end{minipage}
\end{wrapfigure}
The encoder $\mathcal{E}_\phi$ and decoder $\mathcal{D}_\psi$ are implemented as fully-connected neural networks. The encoder  outputs the embedding $z^{i+1}$ given tuples $(s, a, s', r)$ sampled from interaction $\tau^i$, while the decoder reconstructs the transitions and rewards from $\tau^{i+1}$ given the embedding, states, and actions taken. For the RL module, we use soft actor-critic (SAC)~\cite{haarnoja2018soft} as the base algorithm. The actor $\pi$ and critic $Q$ are both fully-connected networks that are conditioned on state $s$ and embedding $z$, and trained with losses $\mathcal{J}_\pi$ and $\mathcal{J}_Q$ from SAC. The pseudocode for our proposed method, Learning and Influencing Latent Intent (LILI), is provided in Algorithm~\ref{alg:latent_strategies}. Additional details can also be found in Appendix~\ref{app:implementation}.

\section{Experiments}
\label{sec:experiments}

A key advantage of our approach is that it enables the ego agent to connect low-level experiences to high-level representations. However, it is not yet clear whether the ego agent can simultaneously learn these high-level representations and intelligently leverage them for decision making. Thus, we here compare our approach to state-of-the-art learning-based methods, some of which also utilize latent representations. We focus on whether the ego agent learns to influence the other agent, and whether this influence leads to greater overall performance. Videos of our results are in the supplementary video and our project webpage: \href{https://sites.google.com/view/latent-strategies/}{\textcolor{orange}{https://sites.google.com/view/latent-strategies/}}.

We select four different learning-based methods that cover a \textit{spectrum} of latent representations:
\begin{itemize}[leftmargin=*]
\vspace{-0.5em}
\setlength\itemsep{0.1em}
    \item \textbf{SAC \cite{haarnoja2018soft}.} At one extreme is soft actor critic (SAC), where the ego agent does not learn any latent representation. This is equivalent to our Algorithm~\ref{alg:latent_strategies} without latent strategies.
    \item \textbf{SLAC \cite{lee2019stochastic}.} In the middle of the spectrum is stochastic latent actor-critic (SLAC), which learns latent representations for a partially observable Markov decision process (POMDP). Within this formulation the other agent's latent strategy could be captured by the POMDP's hidden state.
    \item \textbf{LILAC \cite{xie2020deep}.} Similar to our approach, lifelong latent actor-critic (LILAC) learns a latent representation of the ego agent's environment. However, the latent dynamics in LILAC are only influenced by the environment and unaffected by the ego agent's behavior, such that $z^{i+1} \sim f( \cdot~|~z^i)$.
    \item \textbf{Oracle.} At the other extreme is an oracle, which knows the other agent's hidden intention.
\vspace{-0.5em}
\end{itemize}
We also consider two versions of our approach (Algorithm~\ref{alg:latent_strategies}). \textbf{LILI (No Influence)} is a simplified variant where the ego agent is trained to make greedy decisions that maximize its expected return for the \emph{current} interaction, without considering how these decisions may influence the other agent's future behavior. Our full approach is referred to as \textbf{LILI}, where the ego agent tries to maximize its discounted sum of rewards \emph{across} multiple interactions while taking into account how its own actions may influence the other agent's downstream policy.

\begin{figure}
\centering
\includegraphics[width=.95\columnwidth]{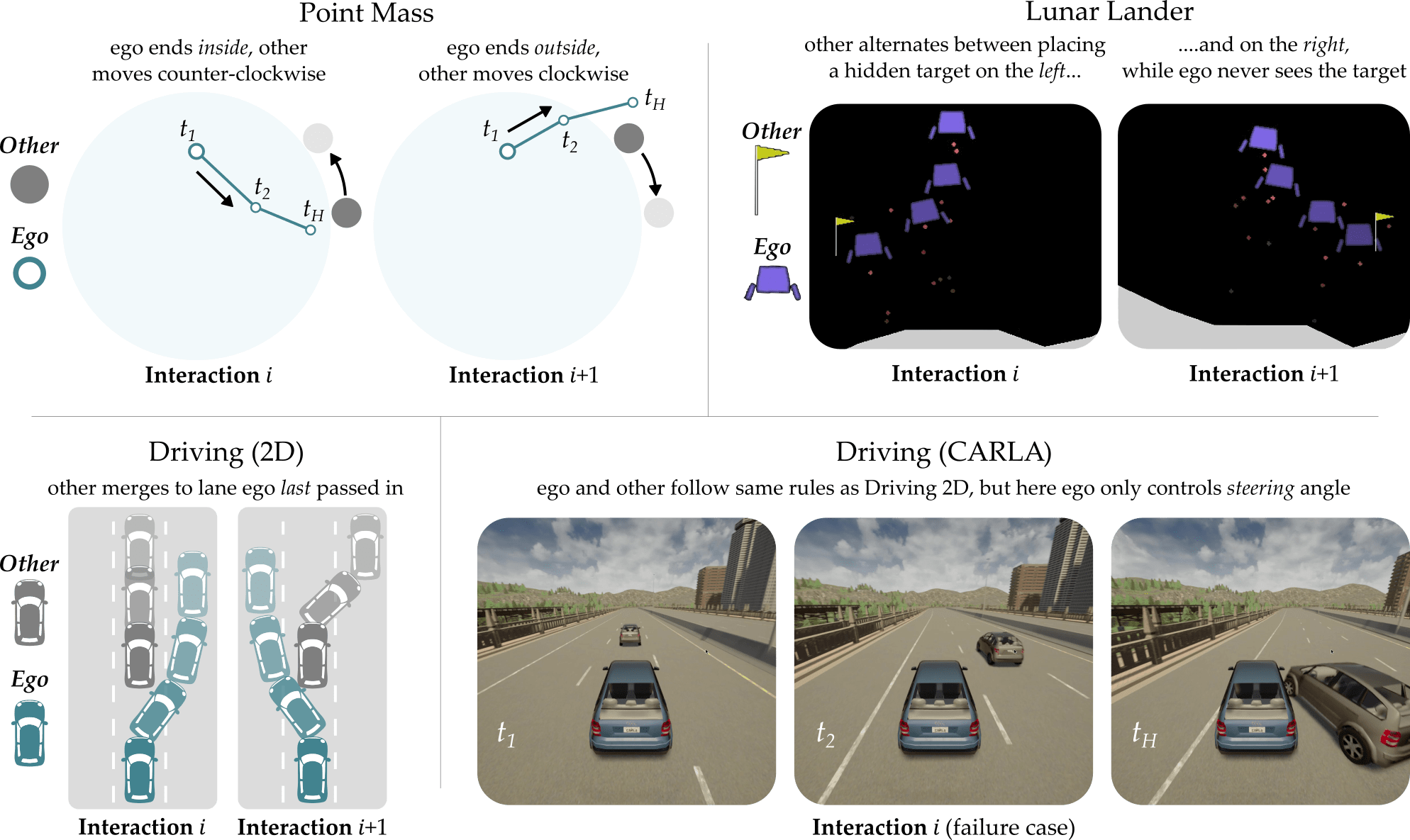}
\vspace{-.5em}
\caption{\small{Simulated environments where the ego agent learns alongside another non-stationary agent. Between interactions, this other agent updates its policy: e.g., moving a hidden target or switching the lane it will merge into. Our approach learns the high-level strategies guiding these policies so both agents seamlessly co-adapt.}}
\label{fig:sim_tasks}
\vspace{-1.2em}
\end{figure}

\subsection{Simulated Environments and Multi-Agent Experimental Setup}

Our simulated environments are visualized in Fig.~\ref{fig:sim_tasks}, and our real robot experiment is shown in Fig.~\ref{fig:front}. All environments have continuous state-action spaces, with additional details in Appendix~\ref{app:experiment_sim}.

\textbf{Point Mass.} Similar to pursuit-evasion games \cite{vidal2002probabilistic}, the ego agent is trying to reach the other agent (i.e., the \textit{target}) in a 2D plane. This target moves one step clockwise or counterclockwise around a circle depending on where the ego agent ended the previous interaction (see Fig.~\ref{fig:sim_tasks}). Importantly, the ego agent never observes the location of the target! Furthermore, because the ego agent starts off-center, some target locations can be reached more efficiently than others.

\textbf{Lunar Lander.} A modified version of the continuous Lunar Lander environment from OpenAI Gym \cite{brockman2016openai}. The ego agent is the lander, and the other agent picks the target it should reach. Similar to point mass, the ego agent never observes the location of the target, 
and must infer the target position from rewards it receives
--- but unlike point mass, the latent dynamics here do not depend on $\tau^i$.

\textbf{Driving (2D and CARLA).} The ego agent is trying to pass another driver. This other agent recognizes that the lane in which the ego agent passes is faster --- and will suddenly merge into this lane during the next interaction. Hence, the ego agent must anticipate how the other car will drive to avoid a collision. We performed two separate simulations with these rules. In Driving (2D), we use a simple driving environment where the ego agent controls its lateral displacement. In Driving (CARLA), we use a 3D simulation where the ego agent must control the vehicle steering \cite{dosovitskiy2017carla}.

\textbf{Air Hockey.} A real-world air hockey game played between two 7-DoF robots (and later between one robot and human). The ego agent is the \emph{blocker}, while the other agent is the \emph{striker}. The striker updates their policy to shoot away from where the ego agent last blocked. When blocking, the ego robot does not know where the striker is aiming, and only observes the vertical position of the puck. We give the ego robot a bonus reward if it blocks a shot on the left of the board --- which should encourage the ego agent to influence the striker into aiming left. See Appendix~\ref{app:experiment_hockey} for more info.

\subsection{Simulation Results: Trapping the Other Agent in Point Mass}

\begin{figure}
\centering
\includegraphics[width=.95\columnwidth]{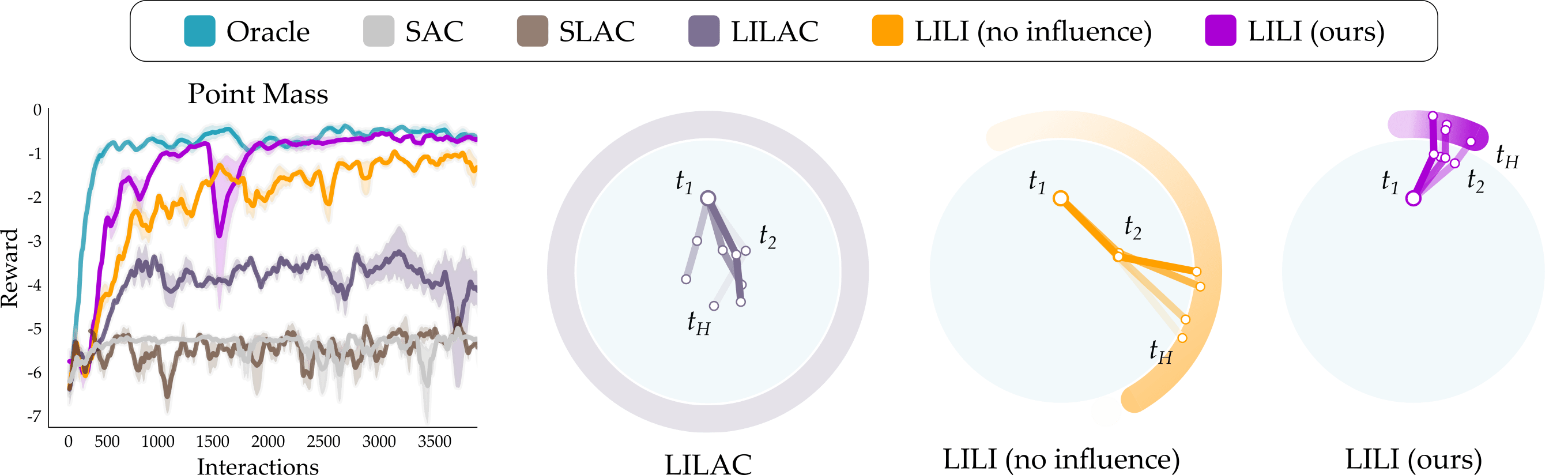}
\vspace{-.6em}
\caption{\small{Results for Point Mass. (Left) Reward that the ego agent receives at each interaction while learning. (Right) Heatmap of the target position during the final $500$ interactions. The ego agent causes the target to move clockwise or counterclockwise by ending the interaction in-or-out of the circle. Our approach \textbf{LILI} exploits these latent dynamics to trap the target close to the start location, decreasing the distance it must travel.}}
\label{fig:sim_results1}
\end{figure}

\begin{figure}
\vspace{-.6em}
\centering
\includegraphics[width=.95\columnwidth]{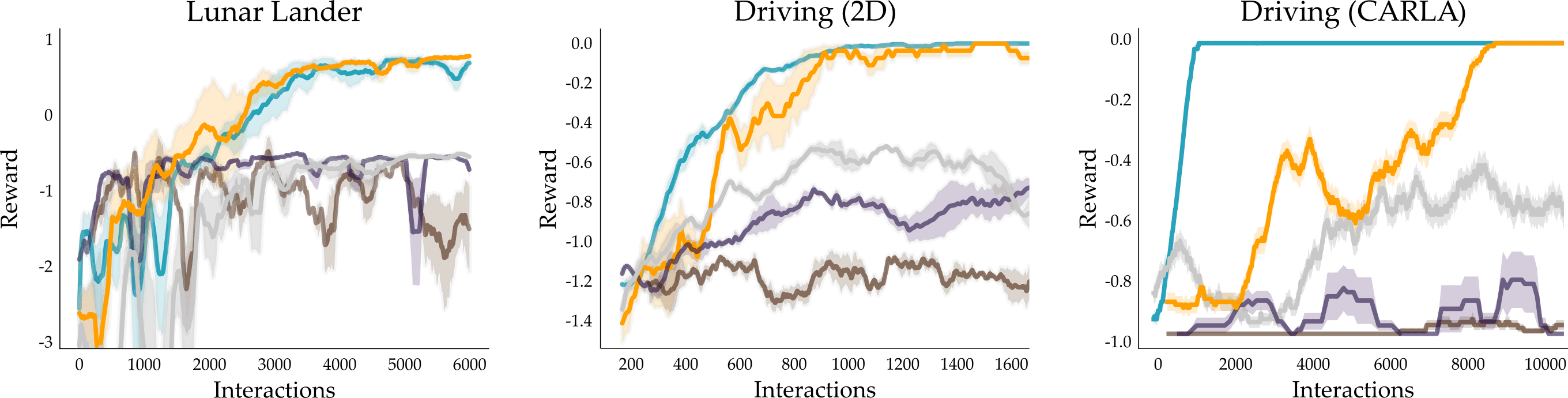}
\vspace{-.6em}
\caption{\small{Results for other simulations. Because each latent strategy was equally useful to the ego agent, here \textbf{LILI (no influence)} is the same as \textbf{LILI}. Shaded regions show standard error of the mean.}}
\label{fig:sim_results2}
\vspace{-1.5em}
\end{figure}

The simulation results are summarized in Figs.~\ref{fig:sim_results1} and \ref{fig:sim_results2}. Across all domains, the proposed method achieves higher returns than existing RL-based algorithms, and nearly matches oracle performance. 

\textbf{Point Mass.} To better explain these results, we specifically focus on the Point Mass domain (Fig.~\ref{fig:sim_results1}). During each timestep the ego agent incurs a cost for its squared Euclidean distance from the hidden target. If the ego agent is \textit{confused} by the changing environment --- and unable to model the target dynamics --- a safe play is simply to move to the center of the circle. Indeed, this behavior is exactly what we see in \textbf{LILAC}, the best-performing baseline. But what if the ego agent successfully models the latent dynamics? A \textit{greedy} ego agent will try to get as close as it can to the target, without considering if it ends the interaction inside or outside of the circle. We observed this na\"ive pursuit in \textbf{LILI (no influence)}, where the ego agent accurately reached the target at each interaction. However, because the ego-agent starts closer to some target locations than others, precisely reaching the target is not always the best long-term plan; instead, it is better to purposely over- or under-shoot, influencing the target towards the start location. As depicted, \textbf{LILI} converges to this final policy that \textit{traps} the other agent. Here the ego agent exploits the latent dynamics, intentionally moving in-and-out of the circle to cause the target to oscillate near the start. Accordingly, \textbf{LILI} achieves higher rewards than \textbf{LILI (no influence)}, and almost matches the gold standard of \textbf{Oracle}.

\textbf{Other Simulations.} Learning results for our remaining domains are shown in Fig.~\ref{fig:sim_results2}. We emphasize that --- because each of the other agent's strategies in these settings is equally beneficial for the ego agent --- here \textbf{LILI} results in the same performance as \textbf{LILI (no influence)}. Further analysis of these results is available in Appendix~\ref{app:behaviors}. 

\textbf{Complex Strategies.} 
So far the other agent's strategy has been relatively straightforward. 
To better understand the robustness of \textbf{LILI}, we next apply our approach to more complex settings.
Specifically, we introduce two different factors to the other agent's strategy: i) \textit{noisy} latent dynamics and ii) latent dynamics that depend on \textit{multiple} previous interactions.
Our results, presented in Appendix~\ref{app:complex_strategies}, demonstrate that \textbf{LILI} is robust to significant levels of noise and is capable of modeling (and influencing) non-Markovian strategies that depend on reasonably long interaction histories.

\begin{figure}[t]
\includegraphics[width=.95\columnwidth]{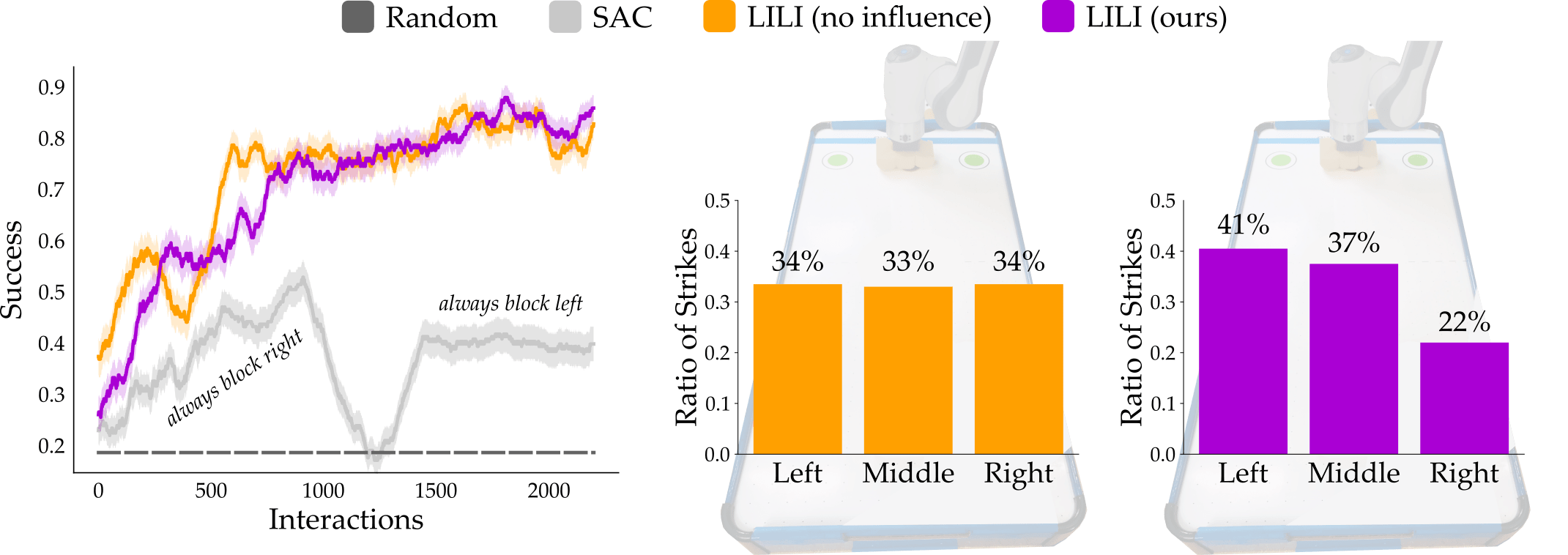}
\vspace{-.8em}
\centering
\caption{\small{Learning results for the air hockey experiment. (Left) Success rate across interactions. (Right) How frequently the opponent fired left, middle, or right during the final $200$ interactions. Because the ego agent receives a bonus reward for blocking left, it should influence the opponent into firing left. At convergence, \textbf{LILI (no influence)} gets an average reward of $1.0 \pm 0.05$ per interaction, while \textbf{LILI} gets $1.15 \pm 0.05$.}}
\label{fig:robot}
\vspace{-1.25em}
\end{figure}

\subsection{Air Hockey Results: Learning to Play Against Robot and Human Opponents}

Building on our simulation results, we perform a real-robot hockey experiment (see Supplementary Video). The ego agent \textit{learns} alongside a robot opponent, and then \textit{plays} against a human opponent. Both human and robot update their policy to aim for the ego agent's weak side.

\textbf{Learning with a Robot Opponent.} Our learning results are summarized in Fig.~\ref{fig:robot}. When moving randomly, the ego agent blocks the puck $18\%$ of the time. By contrast, \textbf{SAC} learns to specialize in blocking \textit{one half} of the board --- either right or left --- and converges to a success rate of $44\%$ over the final $100$ interactions. Only our proposed approach is able to co-adapt with all of the opponent's changing intentions: \textbf{LILI} blocks the puck in $91\%$ of the last $100$ interactions.

Next, recall that the ego agent receives a bonus for blocking on the \textit{left} side of the board. \textbf{LILI (no influence)} fails to guide the opponent into taking advantage of this bonus: the distribution over the opponent's strategies is uniform. However, \textbf{LILI} causes the opponent to strike left $41\%$ of the time. By rule, the striker always changes where they fire: so here the ego agent manipulates the striker into alternating between the left and middle strategies (see example in Fig.~\ref{fig:front}).

\textbf{Playing Against a Human Expert.} We also deploy the converged policies for \textbf{SAC} and \textbf{LILI (no influence)} against an expert human. Like the robot striker, this human aims away from where the ego agent blocked during the previous interaction. The human was given $10$ minutes to practice striking the puck, and then we recorded $40$ consecutive interactions. Our proposed approach performs much better when playing alongside the non-stationary human: \textbf{SAC} blocks $45\%$ of shots ($18/40$), while \textbf{LILI (no influence)} blocks $73\%$ ($29/40$). Please see the supplementary video for examples.

\section{Discussion}
\label{sec:discussion}

\textbf{Summary.} We proposed a framework for multi-agent interaction that represents the low-level policies of non-stationary agents with high-level latent strategies, and incorporates these strategies into an RL algorithm. Robots with our approach were able to anticipate how their behavior would affect another agent's latent strategy, and actively influenced that agent for more seamless co-adaptation. 

\textbf{Limitations.} 
We originally intended to test our approach LILI with human opponents, but we found that --- although LILI worked well when playing against another robot --- the learned policy was too brittle to interact alongside humans. Specifically, LILI learned to block the puck with the edges of its paddle, which led to more failures when humans shot imperfectly. Future work will therefore improve the robustness of our algorithm by learning under noisier conditions and fine-tuning the policy on humans. 
Finally, instead of training alongside artificial agents, we also hope to study the human-in-the-loop setting to adapt to the dynamic needs and preferences of people.



\clearpage
\acknowledgments{This work was supported by NSF Award \#1941722 and \#2006388, 
ONR grant N00014-20-1-2675, and JPMorgan Chase \& Co. Any views or opinions expressed herein are solely those of the authors listed, and may differ from the views and opinions expressed by JPMorgan Chase \& Co. or its affiliates. This material is not a product of the Research Department of J.P. Morgan Securities LLC. This material should not be construed as an individual recommendation for any particular client and is not intended as a recommendation of particular securities, financial instruments or strategies for a particular client. This material does not constitute a solicitation or offer in any jurisdiction.
}


{
\setlength{\bibsep}{1.4pt plus 0.3ex}
\small\bibliography{bibtex}  
}

\newpage
\appendix
\part*{Appendix}

\textbf{COVID-19 Effects.} We are fortunate to test our proposed approach on real world robots. However, the extent of our experiments is limited by the current COVID pandemic. We originally planned to conduct a \textit{full-scale} user study, where $10+$ human participants play hockey against our learning agent. This study would focus on how LILI enables a robot to interact alongside humans who change and update their policy in response to that robot. Unfortunately, local restrictions to lab access prevented us from starting with the robots until mid-June. After setting up the hockey environment and developing LILI \& other algorithms, we performed a proof-of-concept pilot study right before the deadline. As described in Section \ref{sec:discussion}, we here found that LILI was not sufficiently robust for an immediate user study. However, we feel that the necessary changes are straightforward --- e.g., training against a noisier robot opponent --- and therefore we believe that this user study will be successful given normal access to robots and human participants.

\section{Implementation Details}
\label{app:implementation}
Our encoder and decoder networks are both multilayer perceptrons with $2$ fully-connected layers of size $128$ each, and we use a latent space size of $8$. For the reinforcement learning module, the policy and critic networks are MLPs with $2$ fully-connected layers of size $256$. During training, the encoder network weights are updated with gradients from both the representation learning objective $\mathcal{J}_\text{rep}$ and the critic loss $\mathcal{J}_Q$ from the soft actor-critic algorithm.

\section{Experimental Details: Simulated Domains}
\label{app:experiment_sim}
\subsection{Point Mass}
In this environment, the ego agent must reach the target position, which is unknown to the ego. Between interactions, the target $z$ moves along a fixed-radius circle. In particular, if the ego agent's position after $50$ timesteps is inside of the circle, then the target moves $0.2$ radians clockwise after the interaction. If the ego agent ends outside of the circle, the target moves $0.2$ radians counterclockwise. The reward function in this environment is in terms of the distance between the ego and target, $\mathcal{R}(s, z) = - \| s - z \|_2$. For the oracle comparison, we augment the observation space with the target position $z$.

\subsection{Lunar Lander}
The ego agent's goal is to land on the launchpad with $2$ controls (main engine and left-right engine), the position $z$ for which is unknown to the ego. The launchpad alternates between $z = [0.4, 0.2]$ and $z = [-0.4, 0.2]$ at every interaction. Hence, the latent dynamics in the environment are independent of the interactions, and evolve only as a function of time. 
At every time-step, the ego agent receives reward, 
\begin{align*}
    \mathcal{R}(s, a, z) &= -\| s_\text{pos} - z \|_2^2 - \| s_\text{vel} \|_2^2 - | s_\theta | \\ 
    &- 0.0015 * (\text{clip}(a_0, 0, 1) + 1) - 0.0003 * \text{clip}(\text{abs}(a_1), 0.5, 1),
\end{align*}
where the $s_\text{pos}$ denotes the lander's $x$- and $y$-position, $s_\text{vel}$ denotes the lander's $x$- and $y$- velocities, and $s_\theta$ is the lander's angle with respect to the ground. When the interaction terminates, the terminal reward is
\begin{align*}
    \mathcal{R}(s, a, z) &= \begin{cases}
        +1 & \text{if}~\| s_\text{pos} - z \|_2 < 0.1 \\
        -\| s_\text{pos} - z \|_2 & \text{otherwise.}
    \end{cases}
\end{align*}
For the oracle, we augment the observation space with the position of the launchpad $z$. 

\subsection{Driving (2D)}
In this simpler driving domain, the ego agent needs to pass the other agent, which is also quickly switching lanes during the interaction. In interaction $i$, the other agent switches to the lane that the ego agent last selected in interaction $i-1$. Switching to the same lane as the other agent may cause a collision, and the ego agent is given a reward of $-1$ if a collision occurs and $0$ otherwise. The ego agent observes its own position and the other agent's position at each time-step, and the interaction ends after $10$ time-steps. The oracle agent is additionally given knowledge of the lane the other agent plans to switch to as a one-hot vector.

\subsection{Driving (CARLA)}
In the CARLA simulations, our ego agent's goal is to pass an opponent who alternates direction of aggressive lane switches in order to block the lane that the ego agent selected in interaction $i-1$. Unlike the simpler Driving (2D) domain, the ego agent is exposed only to steering wheel angles which affect direction and magnitude of velocity and acceleration. Likewise, the complexity of the environment is increased by the inclusion of early termination for collisions or lane boundary violations, causing episode lengths to vary from $5$ to $20$ time-steps. The ego agent is given a reward of $-1$ for any collisions or lane violations and $0$ otherwise. At each time-step, the ego agent observes both the opponent's and its own position, directional velocity, and angular velocity. The oracle agent is given knowledge of the lane the opponent intends to veer towards as a one-hot vector.

\section{Experimental Details: Robotic Air Hockey}
\label{app:experiment_hockey}
In this domain, we set up two Franka Emika Panda robot arms to play air hockey against one another. Specifically, we designate the ego agent as the blocking robot and the other agent as the striking opponent. The striking mode has three modes: aiming to the left, middle, and right, and switches between these modes as described in Table~\ref{tbl:hockey_opponent}.

\begin{table}[h]
\centering
\vspace{0.1cm}
\begin{tabular}{clc}
\hline
\multicolumn{2}{c}{Current Mode} & Next Mode \\ \hline
\multirow{2}{*}{Left} & Left of ego? & Middle \\
 & Right of ego? & Right \\ \hline
\multirow{2}{*}{Middle} & Left of ego? & Left \\
 & Right of ego? & Right \\ \hline
\multirow{2}{*}{Right} & Left of ego? & Center \\
 & Right of ego? & Left \\ \hline
\end{tabular}
\vspace{0.5cm}
\caption{The opponent is designed to aim \emph{away} from the ego agent, based on the ego agent's final position in the previous interaction. For example, if the opponent last shot down the middle and the ego agent moved slightly to the right to block it, then the puck ended up to the \emph{left} of the ego agent and the opponent will select the left mode next.}
\label{tbl:hockey_opponent}
\end{table}

The ego agent controls its lateral displacement to block pucks along a horizontal axis. At every time-step, the ego agent observes its own end-effector position and the vertical position of the puck, and incurs a cost corresponding to the lateral distance of its end-effector to the puck. An interaction terminates when the puck reaches the ego agent; if the ego successfully blocks the puck, it receives a reward of $+1$ and otherwise $0$. If the ego successfully blocks a left shot, it receives an additional reward of $+1$.

\section{Experimental Results: Analysis of Simulation Results}
\label{app:behaviors}

\subsection{Lunar Lander}
In these experiments, we see near identical convergence rates for \textbf{LILI (no influence)} and the \textbf{Oracle} agents, with both successfully learning to land on the launchpad and model the opponent's alternating coordinates. Due to the simple dynamics of the other agent, baseline agents for \textbf{SAC}, \textbf{SLAC}, and \textbf{LILAC} all performed quite similarly, either randomizing directions or consistently choosing the same path, leading to landing rates near 50\% as expected. Notably, \textbf{LILAC} did not perform better than \textbf{SAC} or \textbf{SLAC}, even though this environment adheres to the assumptions of \textbf{LILAC}. This may be due to optimization challenges in \textbf{LILAC} that are not present in our approach, i.e., regularizing the encoder to a learned prior.

\subsection{Driving (2D)}
In these simulated driving interactions, \textbf{LILI (no influence)} converged towards an optimal solution with near equivalent sample efficiency as the \textbf{Oracle}, demonstrating its effectiveness. Both \textbf{SAC} and \textbf{LILAC} agents learned to avoid rear-end collisions with the opponent, attempting to randomize the choice of passing lane, yielding success rates closer towards the 50\% upper bound. In contrast, the \textbf{SLAC} agent consistently attempted to pass from same lane, resulting in persistent failures.

\subsection{Driving (CARLA)}
The increased complexity of the CARLA simulated environment combined with early interaction terminations provided several stages of learning challenges for agents -- avoiding lane boundary violations, rear-end collisions with the opponent vehicle, and successfully predicting opponent lane switches. Training visualizations indicate that the \textbf{Oracle} agent, after successfully learning to avoid lane boundary violations and rear-end collisions, quickly grasped the lane changing mechanisms and converged to an optimal solution. Both \textbf{SLAC} and \textbf{LILAC} agents suffered particularly from convergence towards repeated local minima -- consistently choosing the same passing lane with only minor steering alterations to extend the episode length, but ultimately colliding with the opponent. Next, the \textbf{SAC} agent performed close to the theoretical no-information upper bound by successfully mixing its strategy. By randomizing which lane to pass from, the \textbf{SAC} agent succeeded nearly 50\% of the time.) Lastly, \textbf{LILI (no influence)} eventually converged to an optimal solution, successfully passing on each interaction. We note that during training, \textbf{LILI (no influence)} successfully learned a pattern of alternation, but often failed to recover the correct alternating parity correctly from occasional rear-end collisions, leading to extended series of failed interactions. In contrast to our other simulated environments, \textbf{LILI (no influence)} converged much slower likely as a result of these early terminating episodes increasing the difficulty in inferring the latent strategy dynamics.

\section{Experimental Results: More Complex Strategies}
\label{app:complex_strategies}

\subsection{Agent Strategies with Noise}
We next study the robustness of our approach by evaluating \textbf{LILI} against agents with noisy strategies. To do so, we modify the Point Mass domain: whereas the other agent in the original task took fixed-size steps between interactions, in this experiment, we add a Gaussian noise to the step size $v$, for $\sigma = 0.2v, 0.4v, 0.6v, 0.8v, 1.0v$. 

In Table~\ref{table:noisy_opponents}, we report the final average rewards and standard error obtained from $3$ random seeds for each value of $\sigma$. The $\sigma = 0.0$ case corresponds to the original, noiseless Point Mass setting. While the performance of \textbf{LILI} slightly degrades as the magnitude of noise increases, \textbf{LILI} can still model the other agent’s strategies and significantly outperforms the \textbf{SAC} baseline, even under very noisy conditions. We observe that \textbf{SAC} and the \textbf{Oracle} perform similarly across different noise levels, and report the aggregated results across all runs. 

\begin{table}[]
\hspace{.025\linewidth}
\begin{minipage}[t]{.44\linewidth}
\centering
\begin{tabular*}{\linewidth}{@{}c@{\extracolsep{\fill}}c@{}}
\toprule
\textbf{Method} & \textbf{Final Rewards} \\ \midrule
Ours-0.0 & -0.7206 +/- 0.0061 \\
Ours-0.2 & -0.8563 +/- 0.0539 \\
Ours-0.4 & -1.039 +/- 0.0311 \\
Ours-0.6 & -1.188 +/- 0.0520 \\
Ours-0.8 & -1.726 +/- 0.1744 \\
Ours-1.0 & -1.715 +/- 0.0535 \\ \midrule
SAC & -5.047 +/- 0.0190 \\
Oracle & -0.5434 +/- 0.0605 \\ \bottomrule
\end{tabular*}
\vspace{0.3cm}
\caption{\textbf{LILI} evaluated in the Point Mass domain against agents with varying noise levels in their strategies. We add Gaussian noise to the other agent's step size $\sigma = 0.0v, 0.2v, 0.4v, 0.6v, 0.8v, 1.0v$, with average step size $v$.}
\label{table:noisy_opponents}
\end{minipage}
\hspace{.1\linewidth}
\begin{minipage}[t]{.44\linewidth}
\centering
\begin{tabular*}{\linewidth}{@{}c@{\extracolsep{\fill}}c@{}}
\toprule
\textbf{Method} & \textbf{Final Rewards} \\ \midrule
Ours-1 & -0.7206 +/- 0.0061 \\
Ours-3 & -0.8969 +/- 0.0276 \\
Ours-5 & -1.736 +/- 0.1481 \\
Ours-7 & -3.232 +/- 0.6576 \\
Ours-9 & -3.046 +/- 0.2468 \\
Ours-11 & -4.486 +/- 0.4979 \\ \midrule
SAC & -5.162 +/- 0.0865 \\
Oracle & -0.5605 +/- 0.0361 \\ \bottomrule
\end{tabular*}
\vspace{0.3cm}
\caption{\textbf{LILI} evaluated in the Point Mass domain against agents whose strategies depend on the previous $N$ interactions, for $N = 1, 3, 5, 7, 9, 11$.}
\label{table:history_opponents}
\end{minipage}
\hspace{.025\linewidth}
\end{table}

\subsection{Agent Strategies with History Dependence}
To understand whether \textbf{LILI} can learn to model more complex strategies, we evaluate our approach against agents whose strategies depend on a longer history of interactions. Concretely, we include an experiment in the Point Mass domain in which the other agent’s strategy now depends on the previous $N$ interactions, instead of only the last interaction, for $N = 3, 5, 7, 9, 11$. Specifically, the other agent moves counterclockwise if, in the majority of the last $N$ interactions, the ego agent’s final position is inside of the circle, and moves clockwise otherwise. We modify \textbf{LILI} so that its encoder takes the last $N$ trajectories as input.

We expect \textbf{LILI} to perform worse as $N$ increases, because the opponent’s strategies become more complex and harder to model. However, we would like to emphasize that we generally expect $N$ to be small when interacting with humans due to their bounded rationality~\cite{simon1997models}, and in this work, we follow this assumption, and are motivated by scenarios where the strategy can be inferred from a short history of interactions.

We report the final average rewards and standard error from $3$ random seeds in Table~\ref{table:history_opponents}. The $N = 1$ case corresponds to the original Point Mass setting. The final performance of \textbf{LILI} degrades as the history length $N$ increases and the other agent’s latent strategies become more difficult to model. \textbf{LILI} still performs better than the \textbf{SAC} baseline for smaller values of $N$, but achieves similar rewards as the \textbf{SAC} baseline for $N = 11$. Because the \textbf{SAC} baseline and the \textbf{Oracle} do not model the other agent’s strategies, we find that they perform similarly across different values of $N > 1$, and report the aggregated results across all runs.

\end{document}